\documentclass[runningheads]{llncs}
\usepackage[T1]{fontenc}
\usepackage{amssymb}
\usepackage{graphicx}
\usepackage{booktabs}
\usepackage{hyperref}
\usepackage{multirow}
\begin{document}
\title {A Few Cases Are All You Need: An Empirical Study of Annotation-Efficient LoRA Fine-Tuning of MedSAM3}
%
\author{Sachin Dudda Nagaraju\inst{1}\and 
Bendik Skarre Abrahamsen\inst{1} \and Ashkan Moradi \inst{1}\and
Mattijs Elschot\inst{1,2}}
\titlerunning{A Few Cases Are All You Need}
\authorrunning{Nagaraju et al.}
%
\institute{Department of Circulation and Medical Imaging, Norwegian University of Science and Technology, Trondheim, Norway \and
Central Staff, St. Olavs Hospital, Trondheim University Hospital, Trondheim, Norway\\
\email{\{sachin.d.nagaraju, bendik.s.abrahamsen, ashkan.moradi, mattijs.elschot\}@ntnu.no}}

\authorrunning{Nagaraju et al.}
%
\maketitle              
\begin{abstract}

Medical image segmentation is essential for clinical workflows such as treatment planning and disease assessment. While specialist tools like TotalSegmentator and MRSegmentator achieve strong performance, they require large annotated datasets for training. Medical foundation models offer a promising alternative through 
large-scale pretraining that reduces the annotation burden for 
new tasks, but zero-shot performance remains 
limited. Parameter-efficient adaptation via Low-Rank Adaptation 
(LoRA) enables efficient specialization with few trainable 
parameters, but a key question remains: how many expert-annotated 
cases are needed to achieve clinically useful segmentation 
performance? We address this by adapting MedSAM3 with LoRA for five abdominal organs (liver, kidneys, spleen, gallbladder, and pancreas) in CT and MRI using only 1, 2, 5, and 10 annotated cases, evaluating on 
AMOS22 dataset. With just 10 cases, models achieve performance competitive with 
specialist systems trained on orders of magnitude more data. 
Notably, this includes reliable gallbladder segmentation 
(Dice 0.68 CT, 0.59 MRI) where existing tools fail almost 
completely (Dice $\leq$\,0.0004), while remaining within 
5--10\% of MRSegmentator for liver, kidneys, and spleen using 
over 100$\times$ fewer annotations. Furthermore, external validation on the Whole Heart Segmentation dataset 
shows that the approach extends to cardiac segmentation, a use 
case beyond the scope of TotalSegmentator (MRI) and 
MRSegmentator, achieving competitive left ventricle (LV) performance with only 
10 annotated cases. Training requires only $\sim$3--5\,hours per organ on a single 
GPU, approximately 2--3$\times$ faster than nnU-Net. These findings suggest that ten annotated cases are 
sufficient for clinically useful segmentation, effectively 
reducing bottlenecks for both image annotation and training time.

\end{abstract}

\section{Introduction}

Medical image segmentation is essential for clinical workflows including treatment planning, surgical guidance, disease assessment, and longitudinal monitoring~\cite{ma2024segment}. Training reliable segmentation models still require large datasets with high-quality expert volumetric annotations, which are difficult to obtain in many clinical settings~\cite{wang2021annotation}. Manual annotation is expensive, time-consuming, and difficult to scale across organs, modalities, and clinical sites~\cite{wasserthal2023totalsegmentator}, limiting model development where only a small number of annotated data are available.

Fully supervised specialist systems have achieved strong segmentation performance. TotalSegmentator segments 104 anatomical structures across CT and 
MRI~\cite{wasserthal2023totalsegmentator}, MRSegmentator 
also provides multi-organ segmentation for both CT and 
MRI~\cite{hantze2025segmenting}, and nnU-Net remains a strong supervised baseline through automatic task adaptation~\cite{isensee2021nnu}. However, all these methods depend on large curated datasets, restricting deployment in low-data settings such as rare diseases, new imaging protocols, or institutions with limited annotation capacity.

Medical foundation models offer a promising alternative. Building on the Segment Anything Model~\cite{kirillov2023segment}, promptable medical segmentation models including MedSAM, SAM-Med3D, and MedSAM3 leverage large-scale pretraining for improved cross-organ generalization~\cite{ma2024segment,wang2025sam,liu2025medsam3}. However, zero-shot performance remains unreliable for small or low-contrast structures such as the gallbladder and pancreas~\cite{wang2025sam}, and full fine-tuning is computationally expensive and prone to overfitting under limited annotation~\cite{wu2025medical,hu2022lora}. Parameter-efficient fine-tuning via Low-Rank Adaptation (LoRA) addresses both concerns by updating only a small fraction of parameters while keeping pretrained weights fixed~\cite{hu2022lora}. Yet an important question remains insufficiently studied: \textit{how many expert-annotated cases are needed to adapt a medical foundation model for clinically useful organ segmentation?} Existing studies report results at fixed annotation budgets or limited task sets, without systematically examining how performance scales with annotation~\cite{wu2025medical,li2024adapting}.

To address this, we conduct a systematic annotation-efficiency 
study of MedSAM3 adapted with LoRA across five abdominal organs 
(liver, kidneys, spleen, gallbladder, and pancreas) in CT and 
MRI, training with 1, 2, 5, and 10 annotated cases and 
evaluating on the independent AMOS22 benchmark~\cite{ji2022amos}. 
We compare against zero-shot foundation models, TotalSegmentator, 
MRSegmentator, and nnU-Net. Cross-center cardiac experiments on 
the WHS dataset~\cite{GAO2023BayeSeg} were also conducted to assess generalization to a use case beyond the scope of the specialist systems. Results show that performance saturates at around 10 annotated 
cases, at which point our models are competitive with specialist 
systems trained on orders of magnitude more data, while training 
2--3$\times$ faster than nnU-Net with only 2.15\% of MedSAM3 parameters 
updated.
 
The main contributions of this work are:
\begin{itemize}
    \item A systematic empirical analysis of annotation efficiency for LoRA-based adaptation of MedSAM3 across abdominal organs and imaging modalities.
    \item Characterization of the performance, and annotation trade-off, showing competitive accuracy already at 10 cases against fully supervised specialist systems with over 100$\times$ fewer annotations.
    \item Cross-center cardiac segmentation experiments on the WHS 
dataset examine generalisability beyond the tasks covered by 
the supervised specialist systems.
\end{itemize}

\section{Related Works}

Deep learning (DL) has greatly improved medical image segmentation. nnU-Net is a widely used baseline because it automatically adapts its preprocessing, architecture, and training pipeline to a given dataset \cite{isensee2021nnu}. Despite its strong performance, it still requires large expert-annotated datasets and task-specific training. Specialist tools have further advanced automatic multi-organ segmentation. TotalSegmentator was trained on 1,204 CT scans and can segment 104 anatomical structures \cite{wasserthal2023totalsegmentator}. MRSegmentator supports both CT and MRI and was developed using more than 2,600 annotated scans to segment 40 structures \cite{hantze2025segmenting}. Although these systems achieve strong performance, they require substantial annotated data and computational resources, which limits their use for rare structures, new imaging protocols, and settings where only a few expert-labeled cases are available. Reducing this annotation requirement, therefore, remains an important challenge.

\subsection{Foundation Models for Medical Image Segmentation}
Vision foundation models offer an alternative to training segmentation networks from scratch. SAM showed strong transferability through large-scale pretraining, but its zero-shot performance on medical images varies considerably across organs, modalities, and prompting strategies \cite{kirillov2023segment}. MedSAM addressed this limitation through large-scale medical pretraining using more than 1.5 million image--mask pairs \cite{ma2024segment}. Later approaches such as MA-SAM and Medical SAM Adapter incorporated parameter-efficient adaptation and volumetric information for CT and MRI segmentation \cite{wu2025medical}. More recently, MedSAM3 and Medical SAM3 extended SAM3 to medical imaging using large multi-dataset training collections \cite{liu2025medsam3,wang2025sam}. Although these models provide stronger medical representations, existing work mainly focuses on improving generalization and adaptation, while the number of expert-annotated cases needed for effective task-specific specialization remains unclear.

\subsection{Few-Shot and Annotation-Efficient Medical Segmentation}
Few-shot methods aim to adapt segmentation models with limited labelled data. Lightweight SAM adaptation has been shown to remain effective even from a single labelled volume \cite{hu2023efficiently}, and pairing data synthesis with LoRA has proven useful for low-data brain tumour and abdominal CT segmentation \cite{feng2023cheap}. Other work has adapted SAM to several anatomical tasks using only a handful of annotated images \cite{xie2024sam}. Although these studies confirm the potential of few-shot adaptation, their experimental settings differ widely, using 2D slices, dataset fractions, or fixed support sets. As a result, it is still unclear how performance changes with the number of fully annotated patient cases. Recent SAM3-based methods also explore parameter-efficient adaptation, but most focus on specific organs, modalities, or larger training collections. A systematic case-level study across multiple organs and both CT and MRI therefore remains limited.
Overall, prior studies show that foundation models and parameter-efficient fine-tuning can reduce the dependence on large labelled datasets. However, most existing work evaluates fixed few-shot settings, individual modalities, or specific anatomical targets. The relationship between the number of fully annotated patient cases and segmentation performance therefore remains insufficiently studied. This work addresses that gap by systematically adapting MedSAM3 with LoRA using 1, 2, 5, and 10 annotated cases across multiple abdominal organs in both CT and MRI, while comparing against established specialist segmentation systems.

\section{Methodology: Annotation-Efficiency Study Design}

This work evaluates the annotation efficiency of MedSAM3~\cite{liu2025medsam3} 
adapted with LoRA~\cite{hu2022lora}. We do not propose a new 
segmentation architecture. Instead, we study how segmentation 
performance changes when only a small number of expert-annotated 
training cases is available. The experiments are conducted in three 
phases. In the first phase, we perform a development-stage analysis to 
select a stable training configuration, including the LoRA rank 
and number of training epochs, so that the final protocol 
requires no per-task hyperparameter search. In the second phase, we fix the selected protocol and train the final models for 30 epochs without validation-based model selection. The final checkpoints are then evaluated directly on independent test datasets.

\subsection{Data and Few-Shot Setting}

For abdominal segmentation, we consider two modalities, CT and 
MRI, across five organs: the liver, kidneys, spleen, 
gallbladder, and pancreas. For each organ and modality, few-shot training sets are selected from the TotalSegmentator CT and MRI datasets using 1, 2, 5, and 10 annotated cases. Each model is trained as an organ- and modality-specific binary segmentation model. The adapted models are evaluated on the independent AMOS22 benchmark dataset to assess generalization.

Subsequently, to further examine whether the observed annotation-efficiency trend extends beyond abdominal anatomy, we further evaluate cardiac segmentation on the WHS dataset. For this experiment, models are trained using 10 annotated CT and MRI cases from selected centers and tested on cases from held-out centers for segmentation of the left ventricle (LV) and right ventricle (RV).

\subsection{MedSAM3 Adaptation with LoRA}

We adopt MedSAM3 as the pretrained foundation model and adapt it using LoRA. During training, all pretrained model weights are kept frozen, and only the LoRA parameters are updated. For a target weight matrix $W \in \mathbb{R}^{m \times n}$, 
LoRA learns a low-rank update such that
\[
W' = W + BA,
\]
where $B \in \mathbb{R}^{m \times r}$ and 
$A \in \mathbb{R}^{r \times n}$, with rank $r \ll \min(m,n)$. 
This allows efficient task-specific adaptation with a small 
number of trainable parameters.

Following the architecture of MedSAM3~\cite{liu2025medsam3}, 
LoRA is applied to all major components: the vision, text, and 
geometry encoders, the DETR encoder and decoder, and the mask 
decoder, covering the full model depth to maximise adaptation 
capacity. The target organ name is provided as the text prompt, and the model is trained to predict a binary mask for the corresponding structure. We optimize the model using a combined Dice and focal loss to address mask overlap and foreground--background imbalance. Training uses AdamW with weight decay, mixed-precision training, gradient clipping, and a linear warm-up followed by cosine learning-rate decay.

\subsection{Training and Evaluation Protocol}

A development-stage analysis is first conducted to select a stable 
LoRA rank and number of training epochs. We evaluate LoRA ranks 
$r \in \{4, 8, 16\}$ at the 10-case annotation budget and monitor 
validation Dice every five epochs. The validation set consists of 
30 CT and 30 MRI cases held out from the TotalSegmentator dataset, selected to cover a representative range of anatomical variability. It is used only to observe training behaviour and is never used for early stopping, learning-rate scheduling, or checkpoint selection. The AMOS22 test set is fully disjoint from both 
the TotalSegmentator training and validation splits, ensuring 
that all reported results reflect genuine cross-dataset 
generalization.

Based on this analysis, the main final evaluation uses LoRA rank $r=16$, a fixed training duration of 30 epochs, and the 10-case annotation budget. No validation-based model selection is performed in Phase~2. The final checkpoint after 30 epochs is used directly for testing. This protocol is applied consistently across organs and modalities, effectively mitigating 
the need for an annotated validation dataset. Dice similarity coefficient is used as a performance metric and we compare the adapted models against zero-shot MedSAM3, TotalSegmentator, MRSegmentator, and nnU-Net trained with larger annotation budgets. For the 10-case setting, three independent runs are performed for all five organs, and the mean and standard deviation are reported.

\section{Experiments and Results}
We report the experiments in three phases: Phase~1 selects the 
training protocol, Phase~2 evaluates annotation efficiency and 
compares against specialist tools on AMOS22, and Phase~3 
assesses generalisation to cardiac segmentation on the WHS 
dataset.
\subsection{Phase 1: Development-Stage Analysis}
Phase~1 was used to define a fixed training protocol for the 
final evaluation. This phase was not intended to report final 
test performance. Instead, it was used to understand training 
behaviour under limited annotation and to select the LoRA rank 
and number of training epochs.

\begin{figure}[!t]
\centering
\includegraphics[width=\textwidth]{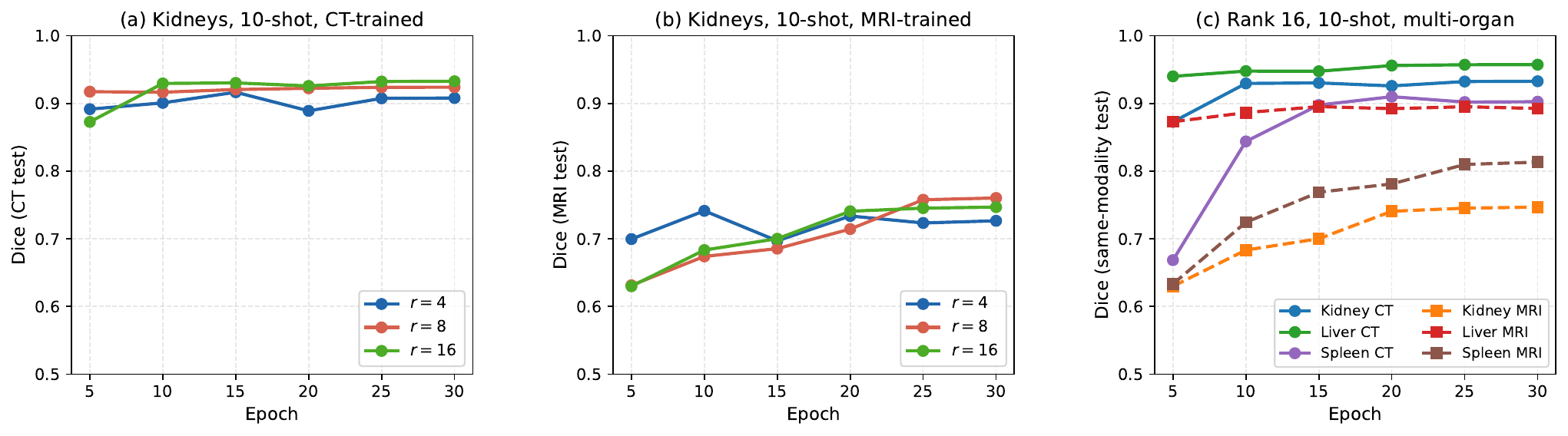}
\caption{Phase~1 development analysis. (a)~Rank comparison 
($r \in \{4,8,16\}$) for kidneys CT-trained, evaluated on CT test 
cases. (b)~Rank comparison for kidneys MRI-trained, evaluated on 
MRI test cases. (c)~Training curves for rank $r{=}16$, 10-shot, 
across three organs and both modalities (solid lines: CT-trained 
evaluated on CT; dashed lines: MRI-trained evaluated on MRI), 
confirming convergence by epoch~30 across all six 
organ/modality combinations.}
\label{fig:phase1}
\end{figure}

We used kidney segmentation as the development task, as it 
represents a moderately challenging organ with clear boundaries 
in both CT and MRI, making it suitable for observing training 
dynamics under limited annotation. We compared LoRA ranks 
$r \in \{4, 8, 16\}$ at the 10-case annotation budget, 
monitoring validation Dice every five epochs. As shown in 
Fig.~\ref{fig:phase1}a and Fig.~\ref{fig:phase1}b, rank $r{=}16$ provides the 
strongest and most stable performance overall across both 
CT-trained and MRI-trained settings, resulting a Dice of 0.93 (CT) and 0.75 (MRI) at epoch 30. 
Therefore, rank $r{=}16$ was selected for all subsequent 
experiments.

To confirm that rank $r{=}16$ generalises beyond kidneys, 
Fig.~\ref{fig:phase1}(c) shows training curves for liver and 
spleen under the same setting. All six organ/modality combinations (kidney, liver, spleen) $\times$ (CT, MRI) plateau in performance by epoch 30, with no signs of overfitting observed in the final epochs. This robustness to overfitting is practically important, as it 
eliminates the need for a validation set and per-organ epoch 
tuning, reducing the annotation burden further beyond the 
10 training cases.

\subsection{Phase 2: Final Evaluation on AMOS22}

Phase~2 evaluates annotation efficiency across all five abdominal 
organs using the fixed protocol (LoRA rank $r{=}16$, 30 epochs). 
Two of these organs (gallbladder and pancreas) were not included 
in the development stage in Phase~1, providing an additional 
test of protocol generalisability. We first examine how performance scales with annotation budget, 
then compare the 10-shot models against specialist tools on the 
independent AMOS22 test set.

\begin{figure}[!t]
\centering
\includegraphics[width=\textwidth]{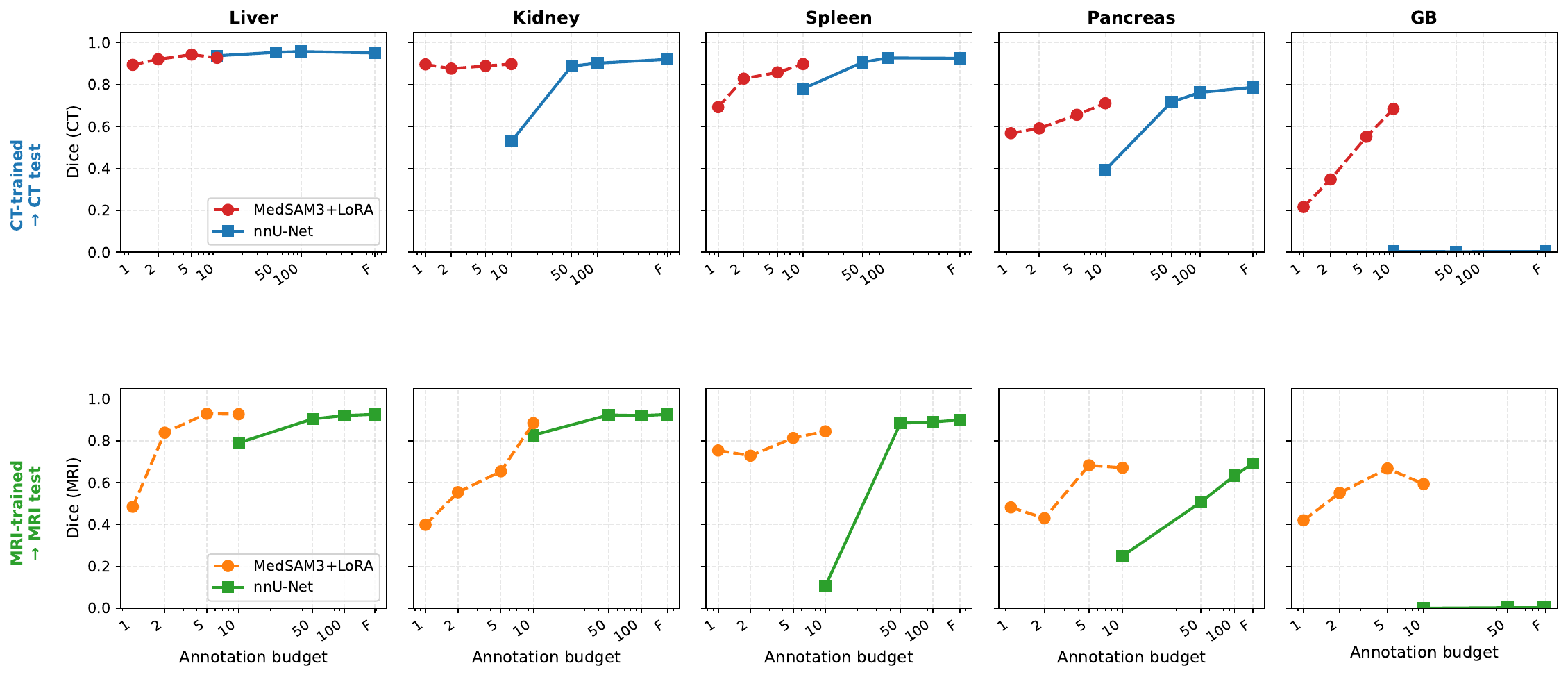}
\caption{Annotation efficiency comparison on AMOS22 test set. 
Top row: CT-trained models evaluated on CT test cases. Bottom 
row: MRI-trained models evaluated on MRI test cases. Dashed 
curves show MedSAM3+LoRA Dice at 1, 2, 5, and 10 annotated 
cases. Solid curves show nnU-Net Dice at 10, 50, 100, and full 
annotation budgets for CT (746, 652, 704, 362, and 491 cases 
for liver, kidney, spleen, pancreas, and gallbladder 
respectively) and 10, 50, 100, and full budgets for MRI 
(193, 175, 180, 146, and 103 cases respectively).}
\label{fig:annotation_budget}
\end{figure}

\paragraph{Annotation efficiency.}
Fig.~\ref{fig:annotation_budget} shows Dice scores for 
MedSAM3+LoRA (1, 2, 5, and 10 annotated cases) and nnU-Net 
(10, 50, 100, and full cases for both CT and MRI, using a 
fixed 80/20 train/validation split with cases randomly 
sampled from the training pool for reduced budgets) across 
all five organs and both modalities. 
MedSAM3+LoRA improves consistently with increasing annotation 
budget, with the largest gains observed between 1 and 5 cases 
and diminishing returns beyond 5 cases, with mean Dice 
improving by 10.6\% (CT) and 24.2\% (MRI) from 1 to 5 cases 
compared to only 4.4\% (CT) and 3.5\% (MRI) from 5 to 10 
cases, confirming saturation at 10 annotated cases. Averaged across all five organs, mean Dice 
improves by 10.6\% (CT) and 24.2\% (MRI) from 1 to 5 cases, 
compared to only 4.4\% (CT) and 3.5\% (MRI) from 5 to 10 
cases, indicating diminishing average performance gains as the annotation budget 
increases.

In contrast, nnU-Net requires substantially 
more annotated cases to reach competitive performance: with only 
10 cases, it degrades severely on MRI spleen (Dice\,=\,0.11) 
and CT pancreas (Dice\,=\,0.39), while MedSAM3+LoRA maintains substantially stronger performance across organs at the same budget. 
This confirms that 10 annotated cases is a practically meaningful 
operating point for foundation model adaptation.

\paragraph{Comparison with specialist tools.}
Table~\ref{tab:amos_main} compares MedSAM3+LoRA at the 10-shot 
operating point against zero-shot foundation models and 
specialist tools. Results are reported as mean\,$\pm$\,std over 
three independent runs. TotalSegmentator~\cite{wasserthal2023totalsegmentator} and 
MRSegmentator~\cite{hantze2025segmenting} used datasets of 
over 1200 annotated scans in total (including training, 
validation, and test splits); our method uses only 10 cases 
per organ and modality. Zero-shot SAM3 and MedSAM3 perform 
poorly across all organs, confirming that prompt-only 
segmentation is insufficient without adaptation. For large 
organs, MedSAM3+LoRA remains within 2--5\% of specialist tools: 
CT-trained achieves 0.928 (liver), 0.898 (kidney), and 0.898 
(spleen); MRI-trained reaches 0.927 (liver) and 0.884 (kidney).

The most notable result is gallbladder segmentation, where both 
specialist tools produce near-zero Dice ($\leq$\,0.0004) on 
AMOS22, likely due to cross-dataset domain shift, organ-size 
variability, and weak contrast. MedSAM3+LoRA achieves 0.684 
(CT) and 0.633 (MRI) with only 10 cases, demonstrating that 
few-shot adaptation recovers structure-specific performance 
where specialist tools fail under dataset shift. For pancreas, 
CT-trained achieves 0.711, surpassing TotalSegmentator (0.624) 
with only 10 cases, though MRSegmentator (0.811 CT, 0.788 MRI) 
remains the stronger specialist for this structure. 


\begin{table}[!t]
\centering
\caption{Dice similarity coefficient on the AMOS22 test set. Each row shows results on CT or MRI test data; columns show CT-trained and MRI-trained models, enabling both same-modality and cross-modality comparison. MedSAM3+LoRA results are mean $\pm$ std over three independent 10-shot runs. Bold: best result per row.}
\label{tab:amos_main}
\scriptsize
\setlength{\tabcolsep}{3.5pt}
\renewcommand{\arraystretch}{1.15}
\resizebox{\textwidth}{!}{%
\begin{tabular}{ll cccc cc}
\toprule
& & \multicolumn{4}{c}{\textit{Specialist / Zero-shot (no fine-tuning)}}
  & \multicolumn{2}{c}{\textit{MedSAM3+LoRA (10-shot)}} \\
\cmidrule(lr){3-6}\cmidrule(lr){7-8}
Organ & Mod.
  & SAM3$_{\textsc{zs}}$ & MedSAM3$_{\textsc{zs}}$ & TotalSeg & MRSeg
  & CT-trained & MRI-trained \\
\midrule
Liver
  & CT  & 0.262 & 0.545 & 0.951 & \textbf{0.959}
         & $0.928\pm0.008$ & $0.881\pm0.081$ \\
  & MRI & 0.453 & 0.331 & 0.921 & \textbf{0.956}
         & $0.918\pm0.008$ & $0.927\pm0.012$ \\
\midrule
Kidney
  & CT  & 0.102 & 0.242 & 0.903 & \textbf{0.925}
         & $0.898\pm0.006$ & $0.864\pm0.005$ \\
  & MRI & 0.122 & 0.140 & 0.875 & \textbf{0.944}
         & $0.852\pm0.022$ & $0.884\pm0.019$ \\
\midrule
Spleen
  & CT  & 0.057 & 0.249 & 0.924 & \textbf{0.952}
         & $0.898\pm0.026$ & $0.786\pm0.005$ \\
  & MRI & 0.153 & 0.111 & 0.712 & \textbf{0.935}
         & $0.900\pm0.021$ & $0.845\pm0.039$ \\
\midrule
Gallbladder
  & CT  & 0.039 & 0.044 & 0.0003 & 0.0004
         & $\mathbf{0.684\pm0.028}$ & $0.561\pm0.044$ \\
  & MRI & 0.030 & 0.013 & 0.0004 & 0.0004
         & $\mathbf{0.633\pm0.034}$ & $0.593\pm0.050$ \\
\midrule
Pancreas
  & CT  & 0.074 & 0.084 & 0.624 & \textbf{0.811}
         & $0.711\pm0.017$ & $0.600\pm0.063$ \\
  & MRI & 0.091 & 0.030 & 0.505 & \textbf{0.788}
         & $0.710\pm0.014$ & $0.671\pm0.017$ \\
\bottomrule
\end{tabular}}
\end{table}

\paragraph{Computational efficiency.}
All experiments were conducted on a single NVIDIA A40 GPU 
(48\,GB VRAM). nnU-Net trains for 500 epochs regardless of annotation budget, 
with a fixed number of gradient steps per epoch independent of 
dataset size, requiring 8.5--11\,GPU-hours per organ. As a 
result, training cost is nearly identical across all annotation 
budgets, offering no computational advantage at low data regimes. MedSAM3+LoRA converges in 30 
epochs, requiring approximately 3--5\,GPU-hours per run, with 
only 2.15\% of parameters updated (rank $r{=}16$; 18.5\,M of 
859\,M total), making it 2--3$\times$ faster than nnU-Net.


Overall, Phase~2 demonstrates that MedSAM3+LoRA with only 10 
annotated cases achieves competitive performance against fully 
supervised specialist tools across all five organs and both 
modalities, while requiring substantially fewer annotations and 
less training time.


\begin{table}[!t]
\centering
\caption{Cross-center cardiac segmentation on WHS. 3D Dice for 
LV and RV under same-modality 10-shot settings, compared against 
TotalSegmentator (CT only) and CARE-WHS 2025 challenge top 
entries trained on 86 fully annotated cases. N/A indicates the 
model does not support the modality or task.}
\label{tab:whs}
\scriptsize
\setlength{\tabcolsep}{4pt}
\renewcommand{\arraystretch}{1.15}
\begin{tabular}{llcccc}
\toprule
Method & Train data & CT LV & CT RV & MRI LV & MRI RV \\
\midrule
TotalSeg~\cite{wasserthal2023totalsegmentator} & large (CT) & 0.896 & 0.915 & N/A & N/A \\
MRSeg~\cite{hantze2025segmenting}              & large      & N/A   & N/A   & N/A & N/A \\
\midrule
MedSAM3+LoRA (CT)  & 10 cases & 0.909 & 0.785 & N/A   & N/A   \\
MedSAM3+LoRA (MRI) & 10 cases & N/A   & N/A   & 0.820 & 0.637 \\
\midrule
CARE-WHS best~\cite{care2025whs} & 86 cases & 0.943 & 0.928 & 0.935 & 0.914 \\
\bottomrule
\end{tabular}
\end{table}

\subsection{Phase 3: Cross-Center Cardiac Segmentation (WHS)}

To assess whether the observed annotation-efficiency trend extends 
beyond abdominal anatomy, and to a use case beyond the scope of 
the supervised specialist systems, we evaluate MedSAM3+LoRA on the WHS 
dataset. CT-trained models use 10 cases from center~A as training data and are tested on 20 held-out cases from center~B to ensure cross center generalizablity.. MRI-trained models use 10 cases 
from center~E and are tested on 20 cases 
from centers~C and~D. Table~\ref{tab:whs} reports same-modality 3D Dice for LV and RV, 
alongside TotalSegmentator (CT only), and the top and median 
entries from the CARE-WHS 2025 challenge trained on all 86 
annotated cases. Specialist system cardiac segmentation was only 
available for TotalSegmentator CT; MRSegmentator and 
TotalSegmentator MRI do not support cardiac structures. CT-trained models achieve Dice of 0.909 (LV) and 
0.785 (RV), surpassing TotalSegmentator CT for LV (0.896) with 
only 10 cases, though falling short for RV (0.915). MRI-trained 
models achieve 0.820 (LV) and 0.637 (RV), demonstrating that 
LoRA adaptation covers a use case where no specialist system is 
available. Overall, MedSAM3+LoRA reasonably generalises to 
cardiac LV and RV segmentation with only 10 annotated cases, 
though performance remains well below the top scores in the 
CARE-WHS challenge, particularly for RV. RV segmentation remains 
more challenging, particularly for MRI, likely due to its more 
variable shape and lower contrast compared to the LV. The 
MRI-trained RV result (Dice\,=\,0.637) reflects the added 
difficulty of cross-center domain shift combined with the 
inherent complexity of RV delineation, where trabeculations 
and thin walls make boundary detection unreliable even for 
fully supervised methods.

\section{Conclusion}
We presented a systematic annotation-efficiency study of 
MedSAM3+LoRA for medical image segmentation. With only 10 
annotated cases, our approach matches specialist tools trained on 
over 100$\times$ more data for large organs, achieves reliable gallbladder segmentation (Dice 0.68 CT, 
0.59 MRI) where the evaluated specialist tools fail almost 
completely (Dice $\leq$\,0.0004), and trains 
2--3$\times$ faster than nnU-Net by updating just 2.15\% of 
parameters. Cross-center experiments on the WHS dataset confirm that the observed annotation-efficiency trend generalizes to a non-abdominal use case, with performance exceeding that of supervised specialist systems. Together, these results 
demonstrate that parameter-efficient fine-tuning of medical 
foundation models with as few as ten expert annotations can be 
sufficient for clinically useful segmentation, substantially 
reducing both the annotation burden and computational cost of 
training task-specific models in low-resource clinical settings.
Future work will extend this framework to more cases of medical use 
and imaging modalities, including federated
learning for privacy-preserving few-shot adaptation across institutions.

\bibliographystyle{splncs04} %
\bibliography{ref}

\end{document}